\documentclass[runningheads]{llncs}
\usepackage{graphicx}
\usepackage{comment}
\usepackage{multirow}
\usepackage{booktabs}
\usepackage{siunitx}
\usepackage{amsmath,amssymb} 
\usepackage{color}
\usepackage{wrapfig}
\usepackage{adjustbox}
\usepackage{textcomp}
\usepackage[T1]{fontenc}

\begin{document}
\pagestyle{headings}
\mainmatter

\title{COCO + LVIS Joint Recognition Challenge Workshop at ECCV 2020 \\
  \large Balanced Activation for Long-tailed Visual Recognition
}

\author{Jiawei Ren\thanks{equal contribution} \and
Cunjun Yu$^\star$\and Zhongang Cai$^\star$\and
Haiyu Zhao
}
\institute{Innova}
\titlerunning{Balanced Activation for Long-tailed Visual Recognition}
\authorrunning{Ren J., Yu C., Cai Z., Zhao H.}

\maketitle

\begin{abstract}
Deep classifiers have achieved great success in visual recognition. However, real-world data is long-tailed by nature, leading to the mismatch between training and testing distributions. In this report, we introduce Balanced Activation (Balanced Softmax and Balanced Sigmoid), an elegant unbiased, and simple extension of Sigmoid and Softmax activation function, to accommodate the label distribution shift between training and testing in object detection. We derive the generalization bound for multiclass Softmax regression and show our loss minimizes the bound. In our experiments, we demonstrate that Balanced Activation generally provides $\sim$3$\%$ gain in terms of mAP on LVIS-1.0 and outperforms the current state-of-the-art methods without introducing any extra parameters. 
\end{abstract}

\section{Introduction}
Most real-world data comes with a long-tailed nature: a few high-frequency classes (or head classes) contribute to most of the observations, while a large number of low-frequency classes (or tail classes) are under-represented in data. In practice, the number of samples per class generally decreases from head to tail classes exponentially. Under the power law, the tails can be undesirably heavy. A model that minimizes empirical risk on long-tailed training datasets often underperforms on a class-balanced test dataset. As datasets are scaling up nowadays, long-tailed nature poses critical difficulties in many vision tasks.

An intuitive solution to long-tailed task is to re-balance the data distribution. Most state-of-the-art (SOTA) methods use the class-balanced sampling or loss re-weighting to ``simulate" a balanced training set~\cite{byrd2018effect,wang2017learning}. However, they may under-represent the head class or have gradient issues during optimization. 

In this report, we first show that the Softmax function is intrinsically biased in the long-tailed scenario. We derive Balanced Activation (Balanced Softmax and Balanced Sigmoid) from the probabilistic perspective that explicitly models the test-time label distribution shift. Theoretically, we found that optimizing for the Balanced Softmax cross-entropy loss is equivalent to minimizing the generalization error bound. We take the off-the-shelf Mask-RCNN as our baseline model and obtain $\sim$3$\%$ mAP improvements on LVIS-1.0 via Balanced Activation without adding any extra computational overhead.

\section{Background}
\textbf{Label Distribution Shift}
\label{sec:label_dist_shift}
We begin by revisiting the multi-class Softmax regression, where we are generally interested in estimating the conditional probability $p(y|x)$, which can be modeled as a multinomial distribution $\phi$:

\begin{equation}
    \phi = \phi_1^{\mathbf{1}\{y=1\}}\phi_2^{\mathbf{1}\{y=2\}} \cdots \phi_k^{\mathbf{1}\{y=k\}};\quad \phi_j = \frac{e^{\eta_j}}{\sum_{i=1}^ke^{\eta_i}};\quad \sum_{j=1}^k\phi_j=1
\end{equation}

\noindent From the Bayesian inference's perspective, $\phi_j$ can also be interpreted as:
\begin{equation}
\phi_j = p(y=j|x) = \frac{p(x|y=j)p(y=j)}{p(x)}
\end{equation}
where $p(y=j)$ is in particular interest under the class-imbalanced setting. Assuming that all instances in the training dataset and the test dataset are generated from the same process $p(x|y=j)$, there could still be a discrepancy between training and testing given different label distribution $p(y=j)$ and evidence $p(x)$. 
With a slight abuse of the notation, we re-define $\phi$ to be the conditional distribution on the balanced test set and define $\hat{\phi}$ to be the conditional probability on the imbalanced training set.
As a result, standard Softmax provides a biased estimation for $\phi$.

\section{Balanced Activation}
To eliminate the discrepancy between the posterior distributions of training and testing, we introduce the Balanced Activation (Balanced Softmax and Balanced Sigmoid). We reach the same conclusion from both the label distribution shift (Section \ref{sec:label_dist_shift}) perspective and the generalization error bound perspective (Section \ref{sec:generalization_error_bound}) that gives rise to Balanced Softmax. The same strategy is used to deduct Balanced Sigmoid. \textbf{We refer readers to Section \ref{sec:tldr} for a summarized version of our findings}. 

\subsection{From Label Distribution Shift Perspective}
We use the same model outputs $\eta$ to parameterize two conditional probabilities: $\phi$ for testing and $\hat{\phi}$ for training. 

\begin{theorem} \label{th:loss}
    Assume $\phi$ to be the desired conditional probability of the balanced data set, with the form $\phi_j={p}(y=j|x)  =\frac{p(x|y=j)}{{p}(x)}\frac{1}{k}$, and $\hat{\phi}$ to be the desired conditional probability of the imbalanced training set, with the form
    $\hat {\phi}_j = {\hat{p}}(y=j|x) = \frac{p(x|y=j)}{\hat{p}(x)}\frac{n_j}{\sum_{i=1}^k n_i}$. If $\phi$
    is expressed by the standard Softmax function of model output $\eta$, then $\hat{\phi}$ can be expressed as
        \begin{equation}\label{eq:our_loss}
            \hat{\phi_j} =\frac{n_je^{\eta_j}}{\sum_{i =1}^k n_ie^{\eta_i}}.
        \end{equation}
\end{theorem}
We use the exponential family parameterization to prove Theorem~\ref{th:loss}. 

\noindent\textbf{Proof.}
Observe that the conditional probability of a categorical distribution can be parameterized as an exponential family. It gives us a standard Softmax function as an \textit{inverse parameter mapping}
\begin{equation}
     \phi_j = \frac{e^{\eta_j}}{\sum_{i=1}^k e^{\eta_i}}
\end{equation}
and also a \textit{canonical link function}:
\begin{equation}\label{eq:link_function}
     \eta_j = \log(\frac{\phi_j}{\phi_k})
\end{equation}

We begin by adding a term $-\log(\phi_j/\hat{\phi}_j)$ to both sides of Eqn. \ref{eq:link_function}, 
\begin{equation}
    \eta_j-\log\frac{\phi_j}{\hat{\phi}_j}=\log(\frac{\phi_j} {\phi_k})-\log(\frac{\phi_j}{\hat{\phi}_j})=\log(\frac{\hat{\phi}_j}{\phi_k})
\end{equation}
Subsequently,
\begin{equation}\label{eq:phi_hat_j}
    \phi_k e^{\eta_j -\log\frac{\phi_j}{\hat{\phi}_j}} = \hat{\phi}_j
\end{equation}
\begin{equation}
    \phi_k \sum_{i=1}^k e^{\eta_i -\log\frac{\phi_i}{\hat{\phi}_i}} = \sum_{i=1}^k \hat{\phi}_i = 1
\end{equation}
\begin{equation}\label{eq:phi_k}
    \phi_k = 1 / \sum_{i=1}^k e^{\eta_i -\log\frac{\phi_i}{\hat{\phi}_i}}
\end{equation}
Substitute Eqn. \ref{eq:phi_k} back to Eqn. \ref{eq:phi_hat_j}, we have
\begin{equation}\label{eq:almost_softmax}
    \hat{\phi}_j = \phi_k e^{\eta_j -\log\frac{\phi_j}{\hat{\phi}_j}} = \frac{e^{\eta_j - \log\frac{\phi_j}{\hat{\phi}_j}}}{\sum_{i=1}^k e^{\eta_i -\log\frac{\phi_i}{\hat{\phi}_i}}}
\end{equation}
Recall that
\begin{equation}
    \phi_j={p}(y=j|x)  =\frac{p(x|y=j)}{{p}(x)}\frac{1}{k} ;\quad
    \hat {\phi}_j = {\hat{p}}(y=j|x) = \frac{p(x|y=j)}{\hat{p}(x)}\frac{n_j}{n}
\end{equation}
then 
\begin{equation} \label{eq:log_phi_hat_phi}
    \log\frac{\phi_j}{\hat{\phi}_j} = \log \frac{n}{kn_j} + \log\frac{\hat{p}(x)}{p(x)}
\end{equation} 
Finally, bring Eqn. \ref{eq:log_phi_hat_phi} back to Eqn. \ref{eq:almost_softmax}
\begin{equation}
    \hat{\phi}_j = \frac{e^{\eta_j - \log \frac{n}{kn_j} - \log\frac{\hat{p}(x)}{p(x)}}}{\sum_{i=1}^k e^{\eta_i -\log \frac{n}{kn_i} - \log\frac{\hat{p}(x)}{p(x)}}} =\frac{n_je^{\eta_j}}{\sum_{i =1}^k n_ie^{\eta_i}}
\end{equation}

Theorem~\ref{th:loss} essentially shows that applying the following Balanced Softmax function can naturally accommodate the label distribution shifts between the training and test sets. We define the Balanced Softmax function as
\begin{equation}\label{eq:our_loss}
\hat{l}(\theta) = -\log(\hat{\phi_y}) =-\log\left(\frac{n_ye^{\eta_y}}{\sum_{i =1}^k n_ie^{\eta_i}}\right).
\end{equation}

\subsection{From Generalization Error Bound Perspective} 
\label{sec:generalization_error_bound}
Generalization error bound gives the upper bound of a model's test error, given its training error. With drastically fewer training samples, the tail classes have much higher generalization error bounds than the head classes, making high classification accuracy on tail classes unlikely. In this section, we show that optimizing Eqn.~\ref{eq:our_loss} is equivalent to minimizing the generalization upper bound.

Margin theory provides a bound based on the margins~\cite{kakade2009complexity}. 
Margin bounds usually negatively correlate to the magnitude of the margin, i.e., a larger margin leads to lower generalization error. Consequently, given a constraint on the sum of margins of all classes, there would be a trade-off between minority classes and majority classes~\cite{LDAM}.

Locating such an optimal margin for multi-class classification is non-trivial. The bound investigated in~\cite{LDAM} was established for binary classification using hinge loss. Here, we try to develop the margin bound for the multi-class Softmax regression. Given the previously defined $\phi$ and $\hat{\phi}$, we derive $\hat{l}(\theta)$ by minimizing the margin bound. Margin bound commonly bounds the 0-1 error:
\begin{equation}\label{eq:01_loss}
    err_{0,1} = \mathrm{Pr}\left[\theta_y^Tf(x)<\max_{i\neq y}\theta_i^Tf(x)\right].
\end{equation}
However, directly using the 0-1 error as the loss function is not ideal for optimization. Instead, negative log likelihood (NLL) is generally considered more suitable. With continuous relaxation of Eqn.~\ref{eq:01_loss}, we have
\begin{equation}
err(t) = \mathrm{Pr}[t<\log(1+\sum_{i\neq y} e^{\theta_i^Tf(x) -\theta_y^Tf(x)})] = \mathrm{Pr}\left[l(\theta)>t\right],
\end{equation}
where $t\geq 0$ is any threshold, and $l_y(\theta)$ is a standard negative log-likelihood with Softmax. This new error is still a counter, but describes how likely the test loss will be larger than a given threshold. Naturally, we define our margin for class $j$ to be
\begin{equation}
    \gamma_j = t - \max_{(x,y)\in S_j} l_j(\theta).
\end{equation}
where $S_j$ is the set of all class $j$ samples. If we force a large margin $\gamma_j$ during training, i.e., force the training loss to be much lower than $t$, then $err(t)$ will be reduced. 
\begin{theorem}
Let $t\geq 0$ be any threshold, for all $\gamma_j >0$, with probability at least $1-\delta$, we have
    \begin{equation}\label{eq:bound}
    err_{bal}(t) \lesssim \frac{1}{k}\sum_{j=1}^k\Big(\frac{1}{\gamma_j}\sqrt{\frac{C}{n_j}}+\frac{\log n}{\sqrt{n_j}}\Big) ; \quad \gamma^*_j = \frac{\beta n_j^{-1/4}}{\sum_{i=1}^{k} n_i^{-1/4}},
    \end{equation}
    where $err_{bal}(t)$ is the error on the balanced test set, $\lesssim$ is used to hide constant terms and $C$ is some measure on complexity. With a constraint on $\sum_{j=1}^k \gamma_j = \beta$, Cauchy-Schwarz inequality gives us the optimal $\gamma^*_j$.
\end{theorem} 

\noindent\textbf{Proof.}
Firstly, we define $f$ as,
\begin{equation}
    f(x) -l(\theta) + t
\end{equation}
where $l(\theta)$ and $t$ is previously defined in submission. However, $f$ does not have a specific semantic meaning as it is defined only to keep consistent with notations in \cite{kakade2009complexity}.

Let $err_{j}(t)$ be the 0-1 loss on example from class $j$
\begin{equation}
    err_{j}(t) = \Pr_{(x,y) \in S_j}[f(x) < 0] = \Pr_{(x,y) \in S_j}[l(\theta)>t] 
\end{equation}
and $err_{\gamma,j}(t)$ be the 0-1 margin loss on example from class $j$
\begin{equation}
    err_{\gamma, j}(t) = \Pr_{(x,y) \in S_j}[f(x) < \gamma_j] = \Pr_{(x,y) \in S_j}[l(\theta) + \gamma_j >t] 
\end{equation}
Let $\hat{err}_{\gamma,j}(t)$ denote the empirical variant of $err_{\gamma,j}(t)$.

For any $\delta > 0$ and with probability at least $1-\delta$, for all $\gamma_j > 0$, and $f \in \mathcal{F}$, Theorem 2 in \cite{kakade2009complexity} directly gives us
\begin{equation}
    err_j(t) \leq \hat{err}_{\gamma, j}(t)+ \frac{4}{\gamma_j}\hat{\mathfrak{R}}_j(\mathcal{F})+\sqrt{\frac{\log(\log_2\frac{4B}{\gamma_j})}{n_j}} + \sqrt{\frac{\log(1/\delta)}{2n_j}}
\end{equation}
where $\sup_{(x,y))\in S}|l(\theta)-t| \leq B$ and $\hat{\mathfrak{R}}_j(\mathcal{F})$ denotes the empirical Rademacher complexity of function family $\mathcal{F}$. By applying \cite{LDAM}'s analysis on the empirical Rademacher complexity and union bound over all classes, we have the generalization error bound for the loss on a  balanced test set
\begin{equation}\label{eq:bound}
    err_{bal}(t) \leq  \frac{1}{k}\sum_{j=1}^k\Big(\hat{err}_{\gamma,j}(t) + \frac{4}{\gamma_j}\sqrt{\frac{C(\mathcal{F})}{n_j}}+\epsilon_j (\gamma_j)\Big)
\end{equation}
where
\begin{equation}
   \epsilon_j (\gamma_j) \triangleq \sqrt{\frac{\log(\log_2\frac{4B}{\gamma_j})}{n_j}} + \sqrt{\frac{\log(1/\delta)}{2n_j}}
\end{equation}
is a low-order term of $n_j$.
To minimize the generalization error bound Eqn. \ref{eq:bound}, we essentially need to minimize

\begin{equation}
    \sum_{j=1}^k \frac{4}{\gamma_j}\sqrt{\frac{C(\mathcal{F})}{n_j}}
\end{equation}

By constraining the sum of $\gamma$ as $\sum_{j=1}^k \gamma_j = \beta$, we can directly apply Cauchy-Schwarz inequality to solve the optimal $\gamma$
\begin{equation}
    \gamma^*_j = \frac{\beta n_j^{-1/4}}{\sum_{i=1}^{k} n_i^{-1/4}}.
\end{equation}

The optimal $\gamma^*$ suggests that we need larger $\gamma$ for the classes with fewer samples. In other words, to achieve the optimal generalization ability, we need to focus on minimizing the training loss of the tail classes. Therefore, for each class $j$, the desired training loss $\hat{l}^*_j(\theta)$ is
\begin{equation}
    \hat{l^*_j}(\theta) = l_j(\theta) + \gamma^*_j, \quad \text{where} \quad l_j(\theta) = -\log (\phi_j),
\end{equation}

\begin{corollary} $\hat{l_j^*}(\theta) =l_j(\theta) +\gamma_j^* =l_j(\theta)+\frac{\beta n_j^{-1/4}}{\sum_{i=1}^{k} n_i^{-1/4}} $ can be approximated by $\hat{l_j}(\theta)$ when:
\begin{equation}
\label{eq:bound_loss}
    \hat{l_j}(\theta) = -\log(\hat{\phi_j});
    \quad \hat{\phi_j} = \frac{e^{\eta_j-\log \gamma_j^*}}{\sum_{i=1}^k e^{\eta_i-\log \gamma_i^*}}= \frac{n_j^{\frac{1}{4}}e^{\eta_j}}{\sum_{i =1}^k n_i^{\frac{1}{4}}e^{\eta_i}}
\end{equation}
\end{corollary}

\noindent\textbf{Proof.}  Notice that 
    $\hat{l_j^*}(\theta) =l_j(\theta) +\gamma_j^* $
 can not be achieved, because$ -\log \hat{\phi_j^*} =-\log \phi_j+\gamma_j^*$ and $\gamma^*_j > 0$ implies
\begin{equation}
    \hat{\phi_j^*} < \phi_j; \quad  \sum_{j=1}^k \hat{\phi_j^*} < \sum_{j=1}^k \phi_j = 1
\end{equation}
The equation above contradicts with the definition that sum of $\hat{\phi^*}$ should be exactly equal to 1.  To solve the contradiction, we introduce a term $\gamma_{base}>0$, such that
\begin{equation}
    -\log \hat{\phi_j^*} =-\log \phi_j - \gamma_{base}+\gamma_j^*; \quad \sum_{j=1}^k \hat{\phi_j^*}=1
\end{equation}
To justify the new term $\gamma_{base}$, we recall the definition of error 
\begin{equation}
    err_{\gamma, j}(t) = \Pr_{(x,y) \in S_j}[l(\theta) + \gamma_j >t]; \quad
    err_{bal}(t) = \Pr_{(x,y) \in S_{bal}}[l(\theta) >t]
\end{equation}
If we tweak the threshold $t$ with the term $\gamma_{base}$
\begin{equation}
    err_{\gamma, j}(t + \gamma_{base}) = \Pr_{(x,y) \in S_j}[l(\theta) + \gamma_j >t+\gamma_{base}] = \Pr_{(x,y) \in S_j}[(l(\theta)-\gamma_{base}) + \gamma_j >t]
\end{equation}
\begin{equation}
    err_{bal}(t + \gamma_{base}) = \Pr_{(x,y) \in S_{bal}}[l(\theta) >t+\gamma_{base}] = \Pr_{(x,y) \in S_{bal}}[(l(\theta)-\gamma_{base})>t]
\end{equation}

As $\gamma^*$ is not a function of $t$, the value of $\gamma^*$ will not be affected by the tweak. Thus, instead of looking for $\hat{l_j^*}(\theta) =l_j(\theta) +\gamma_j^*$ that minimizes the generalization bound for $err_{bal}(t)$, we are in fact looking for $\hat{l_j^*}(\theta) =(l_j(\theta) -\gamma_{base}) +\gamma_j^*$ that minimizes generalization bound for $err_{bal}(t+\gamma_{base})$

Compare the form we obtained from generalization bound Eqn.~\ref{eq:bound_loss} to Eqn.~\ref{eq:our_loss}, we have an additional constant $1/4$ before $\log n_j$. We empirically find that setting $1/4$ to $1$ leads to the optimal results, which may suggest that Eqn.~\ref{eq:bound} is not necessarily tight. To this point, the label distribution shift and generalization bound of multi-class Softmax regression lead us to the same loss form: Eqn.~\ref{eq:our_loss}.

\subsection{Derivation of Balanced Sigmoid}
For instance segmentation, a detector is normally required to predict a background class explicitly. However, the number of background class is uncountable and we need the number of training samples of each class in the training set for Balanced Softmax. Thus, we introduce Balanced Sigmoid to circumvent this problem. By virtue of Bayesian law, a similar strategy can be applied to the multiple binary logistic regressions.

\noindent\textbf{Definition.} Multiple Binary Logisitic Regression uses $k$ binary logistic regression to do multi-class classification. Same as Softmax regression, the predicted label is the class with the maximum model output.
\begin{equation}
    y_{pred} = \arg \max_{j} (\eta_j).
\end{equation}
The only difference is that $\phi_j$ is expressed by a logistic function of $\eta_j$
\begin{equation}
    \phi_j = \frac{e^{\eta_j}}{1+e^{\eta_j}}
\end{equation}
and the loss function sums up binary classification loss on all classes
\begin{equation}
   l(\theta) = \sum_{j=1}^{k}-\log\tilde{\phi_j}
\end{equation}
where
\begin{equation}
    \tilde{\phi_j} =\begin{cases}
            \phi_j, & \text{if $y=j$}\\
            1-\phi_j, & \text{otherwise}
         \end{cases}
\end{equation}

\noindent\textbf{Setup.} By the virtue of Bayes Rule, $\phi_j$  and $1-\phi_j$ can be decomposed as
\begin{equation}\label{eq:bayes_phi_sigmoid}
    \phi_j = \frac{p(x|y=j)p(y=j)}{p(x)} ; \quad
    1-\phi_j = \frac{p(x|y\neq j)p(y\neq j)}{p(x)}
\end{equation}
and for $\hat{\phi}$ and $1-\hat{\phi}$, 
\begin{equation}\label{eq: bayes_phi_hat_sigmoid}
    \hat{\phi}_j = \frac{p(x|y=j)\hat{p}(y=j)}{\hat{p}(x)} ; \quad
    1-\hat{\phi}_j = \frac{p(x|y\neq j)\hat{p}(y\neq j)}{\hat{p}(x)}
\end{equation}

\noindent\textbf{Derivation.} Again, we introduce the exponential family parameterization and have the following link function for $\phi_j$
\begin{equation}
    \eta_j = \log \frac{\phi_j}{1-\phi_j}
\end{equation}
Bring the decomposition Eqn. \ref{eq:bayes_phi_sigmoid}  and Eqn.\ref{eq: bayes_phi_hat_sigmoid} into the link function above
\begin{equation}
    \eta_j = \log (\frac{\hat{\phi}_j}{1-\hat{\phi}_j} \cdot \frac{\phi_j}{\hat{\phi}_j} \cdot \frac{1-\hat{\phi}_j}{1-\phi_j})
\end{equation}
\begin{equation}
    \eta_j = \log (\frac{\hat{\phi}_j}{1-\hat{\phi}_j} \cdot \frac{p(x|y=j)p(y=j)/p(x)}{p(x|y=j)\hat{p}(y=j)/\hat{p}(x)} \cdot \frac{p(x|y \neq j)\hat{p}(y \neq j)/\hat{p}(x)}{p(x|y \neq j)p(y \neq j)/p(x)})
\end{equation}
Simplify the above equation
\begin{equation}
    \eta_j = \log (\frac{\hat{\phi}_j}{1-\hat{\phi}_j} \cdot \frac{p(y=j)}{\hat{p}(y=j)} \cdot \frac{\hat{p}(y \neq j)}{p(y \neq j)})
\end{equation}
Substitute the $n_j$ in to the equation above
\begin{equation}
    \eta_j = \log (\frac{\hat{\phi}_j}{1-\hat{\phi}_j} \cdot \frac{n/k}{n_j} \cdot \frac{n-n_j}{n-n/k})
\end{equation}
Then
\begin{equation}
     \eta_j - \log (\frac{n/k}{n_j} \cdot \frac{n-n_j}{n-n/k}) = \log (\frac{\hat{\phi}_j}{1-\hat{\phi}_j})
\end{equation}
Finally, we have
\begin{equation}
    \hat{\phi}_j = \frac{e^{\eta_j - \log (\frac{n/k}{n_j} \cdot \frac{n-n_j}{n-n/k}) }}{1+e^{\eta_j - \log (\frac{n/k}{n_j} \cdot \frac{n-n_j}{n-n/k})}} 
\end{equation}

\subsection{TL;DR}
\label{sec:tldr}
The label distribution shift and generalization bound of multi-class Softmax regression lead us to the same Balanced Softmax loss form, for certain class $y$, we have:

\begin{equation}
\hat{l}(\theta) = -\log(\hat{\phi_y}) =-\log\left(\frac{n_ye^{\eta_y}}{\sum_{i =1}^k n_ie^{\eta_i}}\right).
\end{equation}

\noindent We also provide the Balanced Sigmoid variant for broader usage:
\begin{equation}
    \hat{l}(\theta)= -\log(\hat{\phi}_y) = -\log\left(\frac{e^{\eta_y - \log (\frac{n/k}{n_y} \cdot \frac{n-n_y}{n-n/k}) }}{1+e^{\eta_y - \log (\frac{n/k}{n_y} \cdot \frac{n-n_y}{n-n/k})}}\right)
\end{equation}

\section{Experiments}
We first validate the proposed method on the long-tailed version of CIFAR-10 and CIFAR-100. Compared to LVIS, these two datasets are relatively small. Hence, they are suitable for fast validation of our ideas. After that, we show that our method leads to a $\sim$3$\%$ improvement over the Repeat Factor Sampling baseline and outperforms current SOTAs by a clear margin on LVIS.

\subsection{CIFAR-10/100-LT}
We show results under different Imbalance Factors in Table \ref{tab:cifar}. The Imbalance Factor is defined as the number of training instances in the largest class divided by that of the smallest. We also visualize the marginal likelihood $p(y)$ on CIFAR-100-LT to show that Balanced Softmax is more stable under a high imbalance factor. We follow \cite{Cui2019ClassBalancedLB} and use ResNet-32 as backbone in all experiments.

\begin{table}[t]
\begin{center}
\fontsize{8}{11}\selectfont
\centering
\begin{tabular*}{\textwidth}{l@{\extracolsep{\fill}}|ccc|ccc} 
\toprule
Dataset & \multicolumn{3}{c|}{CIFAR-10-LT} & \multicolumn{3}{c}{CIFAR-100-LT} \\
\hline
Imbalance Factor & 200 & 100 & 10 & 200 & 100 & 10 \\
\hline
Softmax & 71.2 $\pm$ 0.3  & 77.4 $\pm$ 0.8 & 90.0 $\pm$ 0.2   & 41.0 $\pm$ 0.3 & 45.3 $\pm$ 0.3 & 61.9 $\pm$ 0.1 \\
CBW & 72.5 $\pm$ 0.2 &  78.6 $\pm$ 0.6  & 90.1 $\pm$ 0.2 & 36.7 $\pm$ 0.2 & 42.3 $\pm$ 0.8 & 61.4 $\pm$ 0.3   \\
CBS & 68.3 $\pm$ 0.3& 77.8 $\pm$ 2.2 & 90.2 $\pm$ 0.2 & 37.8 $\pm$0.1  &42.6 $\pm$ 0.4 &61.2 $\pm$ 0.3 \\
Focal Loss~\cite{Lin2017FocalLF} & 71.8 $\pm$ 2.1 & 77.1 $\pm$ 0.2  & 90.3 $\pm$ 0.2   & 40.2 $\pm$ 0.5 &43.8 $\pm$0.1  & 60.0 $\pm$ 0.6 \\
Class Balance Loss~\cite{Cui2019ClassBalancedLB} & 72.6 $\pm$ 1.8 & 78.2 $\pm$ 1.1 & 89.9 $\pm$ 0.3  & 39.9 $\pm$ 0.1 & 44.6 $\pm$ 0.4 & 59.8 $\pm$ 1.1\\
LDAM~\cite{LDAM} & 71.2 $\pm$ 0.3  &  77.2 $\pm$ 0.2 & 90.2 $\pm$ 0.3   & 41.0 $\pm$ 0.3 & 45.4 $\pm$ 0.1 &  62.0 $\pm$ 0.3 \\
Equalization Loss~\cite{Tan2020EqualizationLF} & 72.8 $\pm$ 0.2  & 76.7 $\pm$ 0.1 & 89.9 $\pm$ 0.3 & 43.3 $\pm$ 0.1 & 47.3 $\pm$ 0.1 & 59.7 $\pm$ 0.3 \\
Balanced Softmax & \textbf{79.0} $\pm$ 0.8& \textbf{83.1} $\pm$ 0.4 & \textbf{90.9} $\pm$ 0.4& \textbf{45.9}$\pm$  0.3& \textbf{50.3} $\pm$ 0.3 & \textbf{63.1} $\pm$ 0.2\\
\hline
\end{tabular*}
\end{center}
\caption{
Top 1 accuracy for CIFAR-10/100-LT. Softmax denotes the standard cross-entropy loss with Softmax, CBW denotes class-balanced weighting, and CBS denotes class-balanced sampling. Balanced Activation generally outperforms SOTA methods, especially when the imbalance factor is high. Note that for other methods, we reproduce a higher accuracy than the reported value in the original papers.}
\label{tab:cifar}
\end{table}
\begin{figure*}[!htb]
\vspace{-4mm}
    \centering
    \begin{tabular}{c c c}
    \includegraphics[width=0.33\linewidth]{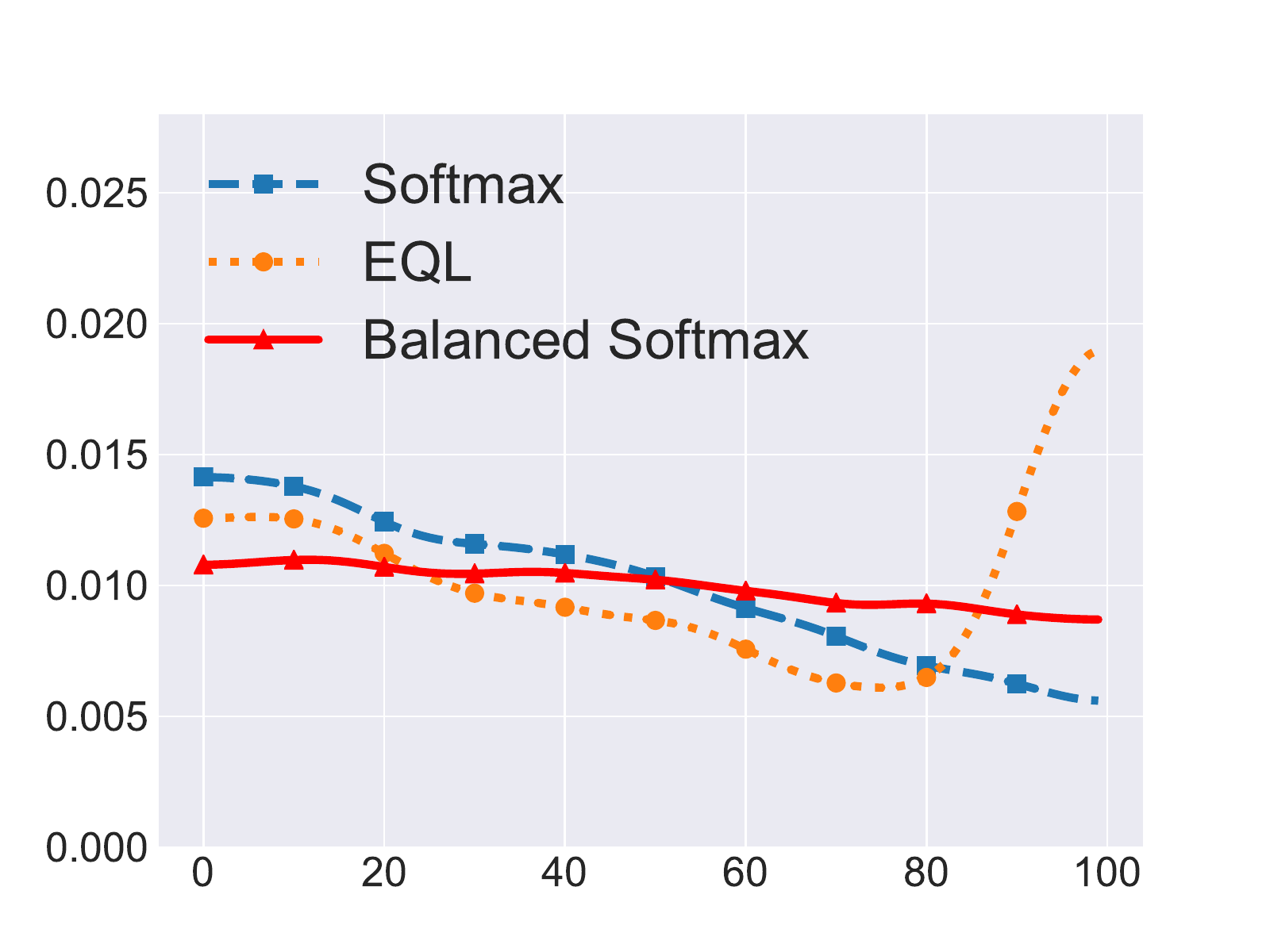}&%
        \includegraphics[width=0.33\linewidth]{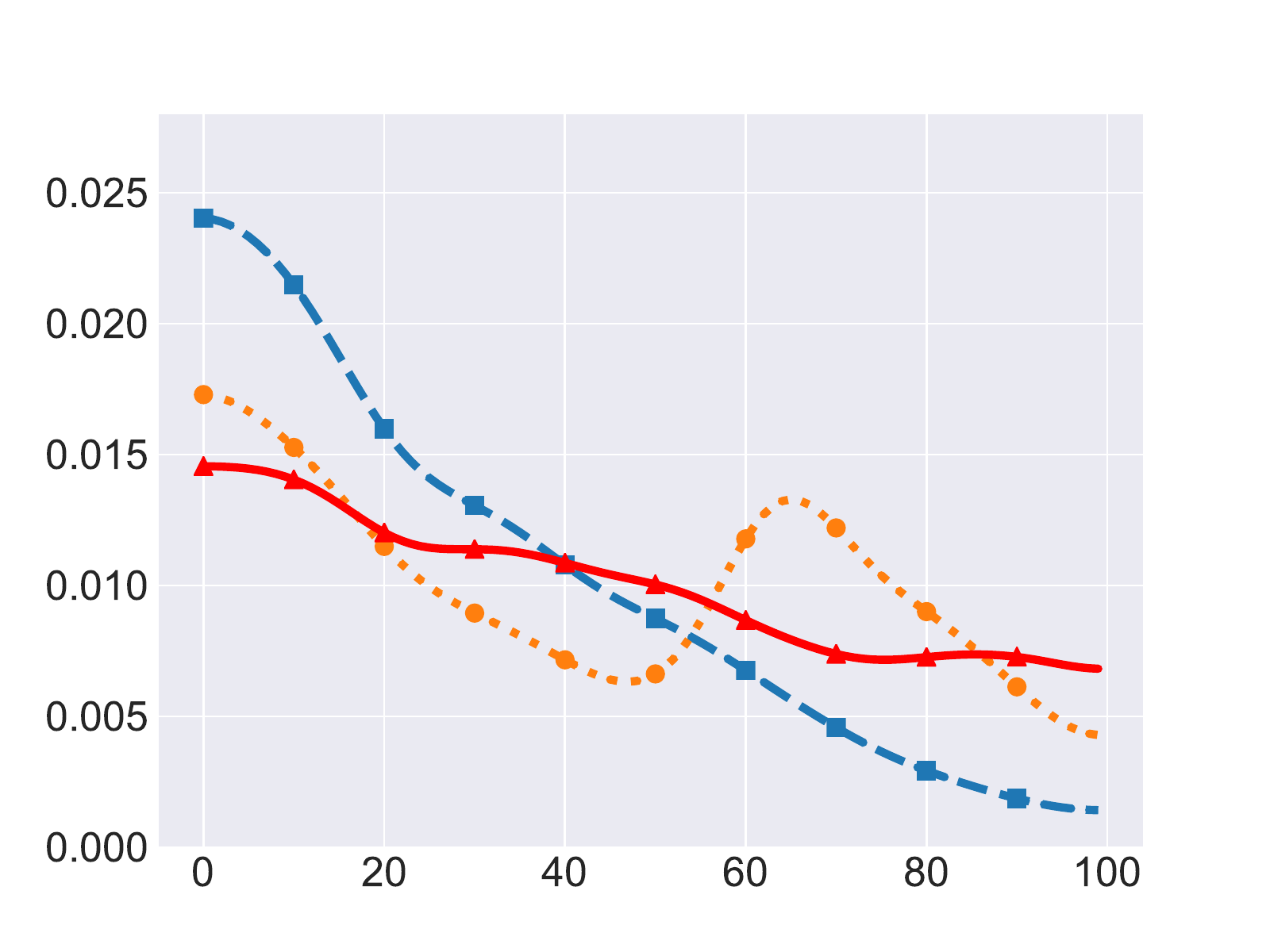}&%
        \includegraphics[width=0.33\linewidth]{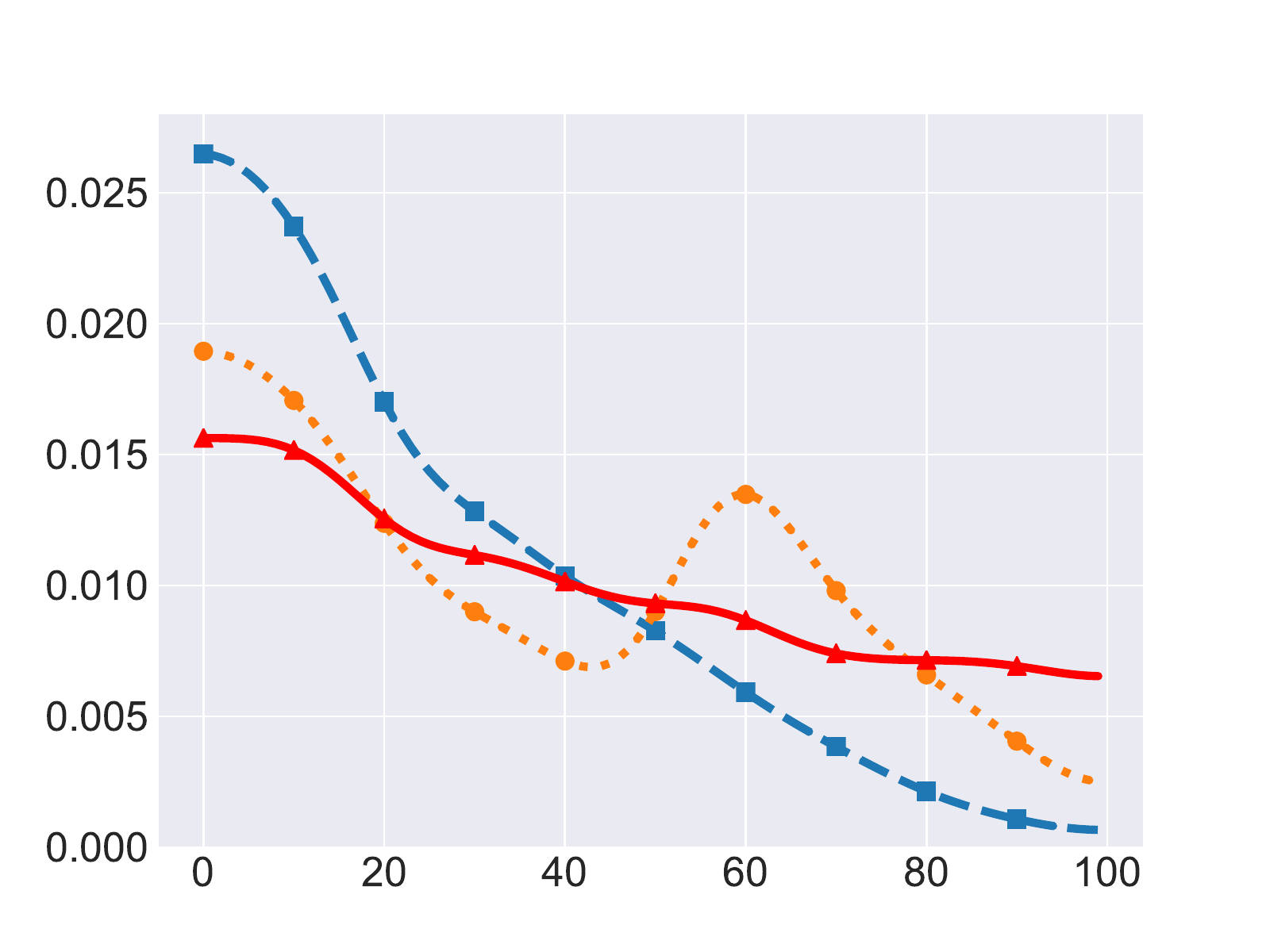}%
    \\  Imbalance Factor = 10 & Imbalance Factor = 100 &Imbalance Factor = 200

    \end{tabular}
    \centering
    \caption{Experiment on CIFAR-100-LT. the x-axis is the class labels with decreasing training samples and the y-axis is the marginal likelihood $p(y)$.
    Balanced Softmax is more stable under a high imbalance factor compared to the Softmax baseline and SOTA method, Equalization Loss (EQL). 
    }
    \label{fig:smooths}
    \vspace{-3mm}
\end{figure*}

\subsection{LVIS-1.0}
For all the experiments on LVIS-1.0, we use the off-the-shelf Mask RCNN and repeat factor sampling as our baseline.We apply scale jitter and random flip at training time (sampling image scale for the shorter side from {640, 672, 704, 736, 768, 800}). For testing, images are resized to a shorter image edge of 800 pixels; no test-time augmentation is used. We train all the repeat factor sampling models for 25 epochs and apply decoupled training for 10 epochs using other methods (cRT, LWS, Equalization Loss and Balanced Sigmoid). Models are all trained on 8 GTX 1080Ti GPUs.
\begin{table*}[t]
\fontsize{8}{11}\selectfont
\begin{center}
    \begin{tabular*}{\textwidth}{l@{\extracolsep{\fill}}|c|c|c|c|c|c|c}
    \hline
    Method & Backbone & \textrm{$AP_{mask}$} & \textrm{$AP_{f}$} &\textrm{$AP_{c}$} &\textrm{$AP_{r}$} &\textrm{$AP_{box}$} & Split \\
    \hline
    \hline
    Repeat Factor Sampling & ResNet-50 &  22.1 & 27.8 & 21.1 & 11.6 & 23.6 & Val \\
    cRT & ResNet-50 & 23.2 & 27.8 & 22.5 & 14.7 & 24.8 & Val \\
    LWS & ResNet-50 & 23.3 & 27.8 & 22.5 & 14.9 & 24.8 & Val \\
    Equalization Loss & ResNet-50 & 24.2 & 27.5 & \textbf{24.2} & 16.5 & 25.8 & Val \\
    Balanced Sigmoid & ResNet-50 &\textbf{25.1}&\textbf{29.3}& 24.0&\textbf{18.3}&\textbf{26.7} & Val\\
    \hline
    Repeat Factor Sampling & ResNet-101 & 23.2 & 28.8 & 22.0 & 13.5 & 24.7 & Val \\
    cRT & ResNet-101 &23.9 &28.7&22.7&15.8&25.6 & Val \\
    LWS & ResNet-101 & 24.2 & 28.8 & 23.1 & 16.5 & 24.7 & Val \\
    Equalization Loss & ResNet-101 & 25.4 & 28.8 & 25.3 & 18.5 & 27.3 & Val \\
    Balanced Sigmoid & ResNet-101 & \textbf{26.4}&\textbf{30.2}&\textbf{25.4}&\textbf{20.1}& \textbf{28.4} & Val   \\
    \hline
    Balanced Sigmoid & ResNet-101 & 26.1 &30.4& 25.0 & 19.5 & - & test-dev   \\
    \hline
    \end{tabular*}
\end{center}
\caption{Comparison of various methods using ResNet-50 and ResNet-101 on LVIS-1.0. Balanced Activation generally outperforms SOTA methods by a healthy margin.}
\label{tab:my_label}
\end{table*}

We compare our method with four existing methods and show improvements:

\noindent\textbf{Repeat Factor Sampling~\cite{gupta2019lvis}:} A simple and effective sampling strategy that increases the rate at which tail categories are observed by oversampling the images that contain them. Different from class-balance sampling, it does not force the sampling probability to be equal for all the classes.

\noindent\textbf{Classifier Re-training (cRT)~\cite{Kang2020DecouplingRA}:} A straightforward approach, which re-train the classifier with class-balanced sampling. The detector is first trained under the instance balance scheme at the first stage. During the second stage, the feature extractor (backbone) is fixed and only the classifier (regression layer) is optimized during training.

\noindent\textbf{Learnable weight scaling (LWS)\cite{Kang2020DecouplingRA}:} An efficient approach to re-balance the decision boundaries of classifiers. The training setup is similar to \textbf{cRT}. The only difference is that there is one more regression layer inserted before the classifier to rebalance the decision boundary. 

\noindent\textbf{Equalization Loss~\cite{Tan2020EqualizationLF}:} The solution of the winner of LVIS Challenge 2019. It pointed out that randomly dropping some scores of tail classes in the Softmax function can effectively help, by balancing the positive gradients and negative gradients flowing through the scoring output.

\section{Conclusion}
We have introduced Balanced Activation (Balanced Sigmoid and Balanced Softmax) for long-tail visual recognition tasks. Balanced Activation tackles the distribution shift between training and testing, combining with generalization error bound theory: it optimizes a Balanced Activation which theoretically minimizes the generalization error bound. We show the method outperforms the current SOTA method without introducing extra parameters. Although Balanced Activation provides a more appropriate objective under the long-tailed scenarios, the optimization process could still be challenging due to the overwhelming gradients from the head classes. A future research direction could be improving the optimization process in complementary to Balanced Activation, for example, a learnable importance sampling method in \cite{Ren2020BalancedMF}.

\clearpage
\bibliographystyle{splncs04}
\bibliography{main}
\end{document}